%% file: la4am.tex
\documentclass[11pt]{article}
\usepackage{style/coling2018}
\usepackage[utf8]{inputenc}
\usepackage{CJKutf8}
\usepackage{times}
\usepackage{latexsym}
\usepackage{graphics}
\usepackage{amssymb,amsmath}
\usepackage{booktabs}
\usepackage{tabularx}
\usepackage{graphicx}
\usepackage{multirow}

\usepackage[vlined, ruled, boxed, linesnumbered]{algorithm2e}
\usepackage{url}
\usepackage{color}

\title{Cross-lingual Argumentation Mining:\\ Machine Translation  (and a bit of Projection) is All You Need!}

\author{%
  Steffen Eger, Johannes Daxenberger, Christian Stab, and Iryna Gurevych\\
  Ubiquitous Knowledge Processing Lab (UKP-TUDA)\\
  Department of Computer Science, Technische Universität Darmstadt\\
  \url{http://www.ukp.tu-darmstadt.de}\\
}

\date{}

\begin{document}
\maketitle
\begin{abstract}
Argumentation mining (AM) requires the identification of complex discourse structures 
and has lately been applied with success monolingually.
In this work, we show that the existing resources are, however, not adequate for assessing cross-lingual AM, due to their heterogeneity or lack of complexity.
We therefore create suitable parallel corpora by 
(human and machine) translating a popular AM dataset consisting of persuasive student essays into German, French, Spanish, and Chinese. 
We then compare (i) annotation projection and (ii) bilingual word embeddings based direct transfer strategies for cross-lingual AM, finding that the former performs considerably better and almost eliminates the loss from cross-lingual transfer.
Moreover, we find that annotation projection works equally well when using either costly human or cheap machine translations. 
Our code and data are available at \url{http://github.com/UKPLab/coling2018-xling_argument_mining}.
\end{abstract}

\blfootnote{
 \hspace{-0.65cm}  
    This work is licensed under a Creative Commons 
    Attribution 4.0 International License.
    License details:
    \url{http://creativecommons.org/licenses/by/4.0/}}

\section{Introduction}\label{sec:introduction}
\input{introduction}

\section{Related Work}\label{sec:related}
\input{related}

\section{Data}\label{sec:data}
\input{data}

\section{Approaches}\label{sec:approach}
\input{approach}

\section{Experiments}\label{sec:experiments}
\input{setup}
\input{experiments}

\section{Conclusion}\label{sec:conclusion}
\input{conclusion}

\section*{Acknowledgements}
This work has been supported by the German Federal Ministry of Education and Research (BMBF) under the promotional reference 01UG1816B (CEDIFOR) and 03VP02540 (ArgumenText). 

\bibliography{coling2018}
\bibliographystyle{style/acl}

\end{document}

%% file: introduction.tex
Argumentation mining (AM) is a fast-growing research field with applications in discourse analysis, summarization, debate modeling, and law, among others \cite{Peldszus2013}.
Recent studies have successfully applied computational methods to analyze monological argumentation \cite{stein:2017j,eger-daxenberger-gurevych:2017:Long}.
Most of these studies view arguments as consisting of (at least) claims and premises---and so do we in this work.
Thereby, our focus is on token-level \emph{argument component extraction}, that is, the segmentation and typing of argument components.

AM has thus far
almost exclusively 
been performed \emph{monolingually}, e.g.\ in English \cite{MochalesPalau2009}, German \cite{EckleKohler2015}, or Chinese \cite{Li2017}.
Working only monolingually is problematic, however, because AM is a difficult task even for humans due to its dependence on background knowledge and parsing of complex pragmatic relations \cite{Moens2017}.
As a result, acquiring (high-quality) datasets for new languages comes at a high cost---be it in terms of training and/or hiring expert annotators or querying large crowds in crowd-sourcing experiments. 
It is thus of utmost importance to train NLP systems in AM that are capable of going cross-language, so that annotation efforts do not have to be multiplied by the number of languages of interest. 
This is in line with current trends in NLP, which increasingly recognize the possibility and the necessity to work cross-lingually, be it in part-of-speech tagging
\cite{Zhang:2016}, dependency parsing \cite{Agic:2016}, sentiment mining \cite{Chen:2016,Zhou:2016}, or other fields. 

In this work, we address the problem of 
cross-lingual (token-level) AM
for the first time.
We initially experiment with available resources in English, German, and Chinese.
We show that the existing datasets for analyzing argumentation are not suitable for assessing cross-lingual component extraction due to their heterogeneity or lack of complexity.
Given this scarcity of homogeneously annotated high-quality
large-scale datasets across different languages, our first contribution is to (1) provide a fully parallel (en-de), \emph{human-translated} version of one of the most popular current AM datasets, namely, the English dataset of persuasive student essays published by \newcite{Stab:2017}. 

We then (2) \emph{machine translate} the 402 student essays into German, Spanish, French, and Chinese. 
Both our human and machine translations contain argumentation annotations, in the form of either human annotations or automatically projected annotations. 
Our experiments 
indicate
that both the translations and the projected annotations are of very high quality, cf.\ examples in 
Table~\ref{table:example}. 

Besides contributing new datasets, (3) we perform the first evaluations of cross-lingual (token-level) AM, based on suitable adaptations of two popular cross-lingual techniques, namely, \emph{direct transfer} \cite{McDonald:2011} and \emph{projection} \cite{Yarowsky:2001}. 
We find that projection works considerably better than direct transfer and almost closes the cross-lingual gap, i.e., cross-lingual performance is almost on par with in-language performance when we use parallel data and project annotations to the target language. This holds both for human (translated, \texttt{HT}) parallel data, which is costly to obtain, and machine translated (\texttt{MT}) parallel data, which is very cheap to obtain for dozens of high-resource languages.

Our findings imply 
that current neural \texttt{MT} has reached a level where it can act as a substitute for costly (non-expert) \texttt{HT} 
even for problems that operate 
on the fine-grained token-level.

%% file: related.tex
In what follows, we briefly summarize the works that most closely relate to our current research. 
\noindent
\paragraph{Argumentation Mining}
AM seeks to automatically identify argument structures in text and has received a lot of attention in NLP lately.
Existing approaches focus, for instance, on specific subtasks like argument unit segmentation \cite{ajjour-EtAl:2017:ArgumentMining}, identifying different types of argument components \cite{MochalesPalau2009,Habernal2017}, recognizing argumentative discourse relations \cite{nguyen-litman:2016:P16-1} or extracting argument components and relations end-to-end \cite{eger-daxenberger-gurevych:2017:Long}. 
However, most of these approaches are specifically designed for English and there are only few resources for other languages.
For German, a few datasets annotated according to the claim-premise scheme are available \cite{EckleKohler2015,Liebeck2016}.
Furthermore, \newcite{Peldszus2016} annotated a small German dataset with claims and 
premises and translated it to English subsequently. 
There are very few works studying AM in other languages, e.g.\ \newcite{Basile2016} for Italian, \newcite{Li2017} for Chinese and \newcite{Sardianos2015} for Greek. There are also two recent papers addressing some form of cross-linguality:
\newcite{Aker:2017} map argumentative sentences from English to Mandarin using machine translation in comparable Wikipedia articles. \newcite{Sliwa:2018} create corpora in Balkan languages and Arabic by labeling the English side of corresponding parallel corpora on the sentence level and then using the same label for the target sentences. In contrast to these works, we work on the more challenging token-level. Moreover, we actually train classifiers for language transfer rather than only creating annotated data in other languages based on parallel data. 

As mentioned, we focus on \emph{cross-lingual component extraction}, that is, the segmentation and typing of argument components in a target language (L2), while having only annotated source language (L1) data. We operate on token-level 
by labeling each token with a BIO label plus its respective component type. The BIO label marks the start, continuation and end of specific argument components.  
Examples 
are given in Tables \ref{table:example} and \ref{table:example-crc}. %

\paragraph{Cross-lingual sequence tagging}
POS tagging and named-entity recognition (NER) are standard tasks in NLP. In recent years, 
there is increased interest 
not only in evaluating POS and NER models within multiple individual languages \cite{Plank:2016}, but also cross-lingually \cite{Zhang:2016,Tsai:2016,Mayhew:2017,Yang:2017}. Two standard approaches are \emph{projection} \cite{Yarowsky:2001,Das:2011,Tackstrom:2013} and \emph{direct transfer} \cite{Tackstrom:2012,Zhang:2016}. Projection uses parallel data to project annotations from one language to another. 
In contrast, in direct transfer, a system is trained in L1 
using 
language independent or 
shared features 
and applied without modification to L2.

While these approaches are typically \emph{unsupervised}, i.e., they assume no annotations in L2, 
there are also \emph{supervised} cross-lingual approaches based on multi-task learning \cite{Cotterell:2017,Yang:2017,Kim:2017}. 
These assume small training sets in L2, 
and a system trained on them is 
regularized by a larger amount of training data in L1. 
In our work, we only consider unsupervised language adaptation, because it is the most realistic cross-lingual scenario for AM, as it may be costly to even produce small amounts of training data in many different languages. 
 
Most cross-lingual sequence tagging approaches address POS tagging, and only few are devoted to NER \cite{Mayhew:2017,Tsai:2016}, aspect-based sentiment classification \cite{lambert:2015}, or even more challenging problems such as discourse parsing \cite{Braud:2017}. While POS tagging and NER are in some sense very similar to AM, namely, insofar as both can be modeled as sequential tagging of tokens, there are also important differences. For example, in POS tagging and NER, the label for a current token usually strongly depends on the token itself plus some local context. This strong association between label, token and local context is largely absent in AM, causing some models that perform well on POS and NER to fail blatantly in AM.\footnote{E.g., we had tried out a word embedding based HMM model \cite{Zhang:2016} in initial experiments but found it to perform below our random baseline. 
The apparent reason is that an HMM cannot deal with long-range dependencies\ that abound in AM.} 

\paragraph{Cross-lingual Word Embeddings}
are the (modern) basis of the direct transfer method. As with monolingual embeddings, there exists a veritable zoo of different approaches, but they often perform very similarly in applications \cite{Upadhyay2016} and seemingly very different approaches are  oftentimes also equivalent on a theoretical level \cite{Ruder:2017}.

%% file: data.tex
\begin{table*}[!tb]
  \centering
  { \small
\begin{tabular}{l|llllllll}\toprule
\bf Name & \bf Docum. & \bf Tokens & \bf Sentences & \bf Major Cl. & \bf Cl. & \bf Prem. & \bf Genre & \bf Lang.\\\midrule
MTX & 112 & 8,865 (en) & 449 & - & 112 & 464 & short texts & en, de \\
CRC & 315 & 21,858 & 957 & 135 & 1,415 & 684 & reviews & zh [en] \\
PE & 402 & 148,186 (en) & 7,141 & 751 & 1,506 & 3,832 & persuasive essays & en [de, fr, es, zh] \\
\bottomrule
\end{tabular}
}
\caption{Statistics for datasets used in this work. Languages in brackets added by the current work.}
\label{table:data}
\end{table*}

\begin{table*}[!tb]
  {\small
  \begin{tabular}{ll}
    Orig-EN & In the end , I think [\textbf{any great success need great work not great luck}] , even though  [\underline{luck is one}
    \\
    & \underline{factor in reaching goal}] but [\textit{its impact is extraneous and we must not reckon on luck in our plans}] .\\
    \texttt{HT}-DE-HumanAnno & Schlie\ss{}lich denke ich , dass [\textbf{jeder gro\ss{}e Erfolg auf harter Arbeit statt Gl\"uck beruht}] , obwohl\\
    & 
[\underline{Gl\"uck ein Faktor in der Erreichung des Ziels ist}] , jedoch [\emph{ist dessen Auswirkung unwesentlich und}\\
   &
\emph{wir sollten uns nicht in unseren Projekten auf unser Gl\"uck verlassen}] .\\
\texttt{HT}-DE-ProjAnno & Schlie\ss{}lich denke ich , dass [\textbf{jeder gro\ss{}e Erfolg auf harter Arbeit statt Gl\"uck beruht , obwohl}\\
    & 
\textbf{Gl\"uck}] [\underline{ein Faktor in der Erreichung des Ziels ist}] , jedoch [\emph{ist dessen Auswirkung unwesentlich und}\\
& \emph{wir sollten uns nicht in unseren Projekten auf unser Gl\"uck verlassen}] .\\
\texttt{MT}-DE-ProjAnno & Am Ende denke ich , dass [\textbf{jeder gro\ss{}e Erfolg gro\ss{}e Arbeit erfordert , nicht viel Gl\"uck}] , auch \\
& wenn [\underline{Gl\"uck ein Faktor beim Erreichen des Ziels}] [\emph{ist , aber seine Auswirkungen sind irrelevant
und}\\
& \emph{wir d\"urfen nicht mit Gl\"uck in unseren Pl\"anen rechnen}] . \\
    \texttt{MT}-ES-ProjAnno & Al final , creo que [\textbf{cualquier gran éxito requiere un gran trabajo y no mucha suerte}] , aunque
la \\
& [\underline{suerte es un factor para alcanzar el objetivo}] , pero [\emph{su impacto es extraño y no debemos tener en
}\\
& \emph{cuenta la suerte en nuestros planes}] .
    \\
\texttt{MT}-FR-ProjAnno & 
En fin de compte , je pense que [\textbf{tout grand succès a besoin d' un bon travail , pas de chance}] ,
\\
& même si la [\underline{chance est un facteur d' atteinte de l' objectif}] , mais [\emph{son impact est étranger et nous ne} \\
& \emph{devons pas compter sur la chance dans nos plans}] .
\\
\texttt{MT}-ZH-ProjAnno & 
\begin{CJK*}{UTF8}{gbsn}
最后 , 我 认为 [\textbf{任何 伟大 的 成功 都 需要 伟大 的 工作 , 而不是 运气 好}] , 即使 \end{CJK*}\\
& \begin{CJK*}{UTF8}{gbsn}
[\underline{运气 是
达成 目标 的 一个 因素}] , [\emph{但 其 影响 是 无关 紧要 的 , 我们 不能 算 计划 中的 运气}] 。
\end{CJK*}
  \end{tabular}
  } 
  \caption{Human-annotated English sentence in the PE dataset as well as translations with human-created and projected annotations. Major claims in bold, claims underlined, premises in italics. \texttt{HT}/\texttt{MT} $=$human/machine translation. }
  \label{table:example}
\end{table*}

We chose three freely available datasets: a small parallel German-English dataset, and considerably larger English and Chinese datasets 
using (almost) the same inventory of argument types, 
which we therefore assumed to be adequate for cross-lingual experiments. We translated the two last named monolingual datasets in other languages, described below. 
Statistics for all datasets are given in Table \ref{table:data}. 

\subsection{Microtexts (MTX)}

\newcite{Peldszus2016} annotated 112 German short texts (six or less sentences) written in response to questions typically phrased like ``Should one do X''. 
These were annotated according to a version of Freeman's theory of argumentation macro-structure \cite{Peldszus2013b}.
Each microtext consists of one (central) claim and several premises.
As opposed to our other datasets, MTX has no ``O'' (non-argumentative) tokens and no major claims.
The German sentences have been professionally translated to English, making this the first parallel corpus for AM in English and German.

\subsection{Chinese Review Corpus (CRC)}
\newcite{Li2017} created the only large-scale argument mining dataset in Chinese,
freely available and with annotations on component level according to the claim-premise scheme \cite{Stab:2017}.
We thus chose to include this 
dataset in our experiments, despite differences in the domain of the annotated texts. 
\newcite{Li2017} used crowdsourcing to annotate Chinese hotel reviews from \textit{tripadvisor.com} with four component types (major claim, claim, premise, premise supporting an implicit claim).
We consider only those components 
with
direct overlap with the components used by \newcite{Stab:2017}, thus considering components labeled as ``premise supporting an implicit claim'' as non-argumentative. 
We applied the CRF-based Chinese word segmenter by \newcite{Tseng2005} to split Chinese character streams into tokens.
Furthermore, we only use the ``Easy Reviews Corpus'' from \newcite{Li2017}. The remaining part of the corpus are isolated sentences from reviews with low overall inter-annotator agreement, which we ignored.
An example from CRC can be found in Table~\ref{table:example-crc}.

\begin{table*}[!tb]
  {\small
  \begin{tabular}{ll}
    Orig-ZH & 
    \begin{CJK*}{UTF8}{gbsn}
    几 次 入住 中国 大 饭店 , [\textbf{感觉 都 非常 不错}] , [\underline{新开 的 豪华 阁 酒廊 非常 棒}] , [\underline{饮料 丰
富}] , \end{CJK*}
\\ & 
\begin{CJK*}{UTF8}{gbsn}
[\underline{食物 也 很 好吃}] , [\underline{服务 也 非常 的 棒}] , [\textit{尤其 特别 感谢 杨雪峰 和 张东静 , 他们
非常 贴心}]
    \end{CJK*}
    \\
    \texttt{MT}-EN-ProjAnno & 
    Several times staying at China World Hotel , [\textbf{I feel very good}] , the [\underline{newly opened Horizon} \\
%
& \underline{Club Lounge is great}] , [\underline{rich drinks}] , [\underline{food is also very good}] , [\underline{very good service}] ,\\
& [\emph{especially thanks
to Yang Xuefeng and Zhang Dongjing , they are very caring}]
    \\
  \end{tabular}
  } 
  \caption{Review from CRC corpus as well as English machine translation with projected annotations. Major claims in bold, claims underlined, premises in italics (ZH: regular font).}
  \label{table:example-crc}
\end{table*}

\subsection{A Large-Scale Parallel Dataset of Persuasive Essays (PE)}
\newcite{Stab:2017} created a dataset of persuasive essays written by students on \textit{essaysforum.com}. 
These are about controversial topics such as ``competition or cooperation---which is better?''.
To obtain a {human-translated} parallel version of this dataset, we asked seven native speakers of German with an attested strong competence in English (all students or university employees) to translate the 402 student essays in the PE corpus sentence-by-sentence.
As only requirement, we asked the translators to retain the argumentative structure in their translations: i.e., the translation of an argument component should be connected and not contain non-argumentative tokens. Since German has a freer word order compared to English, this requirement can in virtually all cases be easily fulfilled without producing awkward sounding German translations. Each essay was translated by exactly one translator. 
Besides translating the essays, we also asked the translators to annotate argument boundaries so that the original mark-up is preserved in the translations. 
The translators took about 40min on average to translate one essay and indicate the argument structures. 
Thus, they required about 270 hours to translate the whole PE corpus into German, and the resulting overall cost 
was roughly 3,000 USD. The motivations to ask translators to translate argument components contiguously were that (i) all monolingual AM datasets we know of have contiguous components, (ii) transfer would have been naturally hampered had components in the source language been contiguous but not in the target language, at least for methods such as direct transfer.\footnote{We note that even  professional translations  typically differ from original, non-translated texts because they retain traces of the source language \cite{Rabinovich:2017}. 
We thus speculate that our reported results are probably slightly upward biased compared to a situation where the test data consists of original German student essays. This latter situation would have been much more costly to produce, in any way: it would have required retrieval (and, if necessary, creation) of original student essays in German as well as induction of all subsequent annotation mark-up.} 

To obtain further parallel versions of the PE data, we also \emph{automatically} translated them into German, French, Spanish, and Chinese using Google Translate.
Of course, we cannot make any demands on how Google Translate translates text into other languages but noticed that it has a tendency to stay rather close to the original text, but nevertheless has a very high perceived quality of translation. We automatically projected argument structures 
from the English text to the machine translations using our projection algorithm described in \S\ref{sec:approach}. It took few hours to automatically translate the PE corpus into the four languages.
Examples of the data as well as the human and machine translations can be found in Table \ref{table:example}. Even though we also provide translations of PE in French, Spanish and Chinese, our primary focus in our experiments below is on the languages for which we have gold (human-created) data, i.e., EN$\leftrightarrow$DE (for PE and MTX) as well as EN$\leftrightarrow$ZH.

%% file: approach.tex
In what follows, we describe our adaptations of direct transfer and projection to the AM task. 
Direct transfer focuses on the source language and trains on human-created L1 data as well as human-created L1 labels. 
In contrast, during training, projection operates directly on the language of interest, viz., L2. 
This comes at a cost: the labels in L2 are noisy, because they are projected from L1, which is an error-prone process. The success of projection can therefore be expected to largely depend on 
the quality of this transfer step. 
Projection makes stronger assumptions than direct transfer: it requires parallel data.\footnote{Thus, direct transfer is potentially the cheaper approach, even though it also requires bilingual word embeddings, which themselves are based on some form of bilingual signal, e.g., parallel sentence- or word-level data.}
When the parallel data is induced via machine translation, then a second source of noise for projection 
is the `unreliable' L2 input training data.

\paragraph{Direct Transfer}
Here, we directly train a system on bilingual representations,
which in our case come in the form of bilingual word embeddings. 
To retain some freedom over the choice and parameters of our word embeddings, we choose to train them ourselves instead of using pre-trained ones. For EN$\leftrightarrow$DE we induce bilingual word embeddings by training BIVCD \cite{Vulic:2015} and BISKIP models \cite{Luong:2015} on $>$2 million aligned sentences from the Europarl corpus \cite{Koehn:2005}. BIVCD 
concatenates bilingually aligned sentences (or documents), randomly shuffles each concatenation and trains a standard monolingual word embedding technique on the result; here, we use the word2vec skip-gram model \cite{Mikolov:2013}. 
BIVCD was shown to be competitive to more challenging approaches in \newcite{Upadhyay2016}. BISKIP is a variant of the standard skip-gram model which predicts mono- and cross-lingual contexts. It requires word alignments between parallel sentences and we use fast-align for this \cite{Dyer:2013}. For EN$\leftrightarrow$ZH we train the same models on the UN corpus \cite{Ziemski:2016}, which comprises $>$11 million parallel sentences. We train embeddings of sizes 100 and 200.

\paragraph{Projection}
To implement projection for the problem of token-level AM, we proceed as
follows. 
We take our human-labeled L1 data and align it with its
corresponding parallel L2 data using 
fast-align. 
Once we have word level
alignment information, we consider for each argument component $c(s)$ in L1
of type $a$ (e.g., MajorClaim, Claim, Premise) with consecutive
words $s_1,\ldots,s_{N}$: the word $t_1$ with smallest index in the corresponding 
L2 sentence that is aligned to some word in $s_1,\ldots,s_N$, and
the analogous word $t_{-1}$ with largest such index.
We then label all the words in the L2 sentence between
$t_1$ and $t_{-1}$ with type $a$, using a correct BIO structure,
resulting in $c(t)$.
We
repeat this process for all the components within a sentence in
L1 and 
for all sentences. In case of collision, e.g., if two components in L2
would overlap according to the above-described strategies, we simply
increment the beginning counter of one of the components until they
are disjoint. If our above strategy fails, i.e., $c(s)$ cannot be
projected, e.g., because the words in an L1 component are not aligned to any words in L2, 
then we
simply ignore the projection of $c(s)$ to the L2 sentence, labeling the corresponding words in L2 as non-argumentative instead. 
We think of this projection strategy as naive because we do not do much to resolve conflicts and instead trust the quality of the alignments and that the subsequent systems trained on the projected data are capable of gracefully recovering from noise in the projections.

%% file: setup.tex
We perform token-level sequence tagging. Our label space is $\mathcal{Y}=\{\text{B,I}\}\times T\:\cup\:\{\text{O}\}$ where $T$ is the set of argument types, comprising ``claim'', ``premise'', and (if applicable) ``major claim''.  

\subsection{Experimental Setup}\label{sec:setup}
To perform token-level sequence tagging, we implement a standard bidirectional LSTM with 
a CRF layer as output layer in TensorFlow.
The CRF layer accounts for dependencies between successive labels.
We represent words by their respective embeddings. 
In addition to this word-based information, we also allow the model to learn a character-based representation (via another LSTM) and concatenate this learned representation to the word embedding. 
Our model is essentially the same as the ones proposed by \newcite{Ma:2016} and \newcite{Lample:2016}; it is also a state-of-the-art model for monolingual AM \cite{eger-daxenberger-gurevych:2017:Long}. We 
name it BLCRF$+$char, when character information is included, and BLCRF when disabled. 
For all experiments, we use the same architectural setup: we use two LSTM hidden layers with 100 hidden units each. 
We train for 50 epochs using a patience of 10.
We apply dropout on the embeddings as well as on the LSTM units. 
On character-level, we also use a bidirectional LSTM with 50 hidden units and learn a representation of size 30. 
As evaluation measure we choose macro-F1 as implemented in scikit-learn \cite{scikit-learn}. 

\paragraph{Baseline} A simple baseline to test successful learning is to choose the majority label in the test data. However, this performs particularly poorly on token-level and for our chosen evaluation metric. We therefore choose a more sophisticated baseline. 
We first split our datasets by sentences and then compute a probability distribution of how likely each argument component appears in a sentence. At test time, we again split the test data by sentences and then label each token in the test sentence with a randomly drawn argument component (according to the calculated probability distribution on train/dev sets). We label all the tokens in the sentence with the drawn argument component type, keeping valid BIO structure. We label the last token (which is typically a punctuation symbol) with the ``O'' label in PE and CRC. 
In essence, our baseline is a random baseline, but has some basic prior knowledge of the BIO format. 

\paragraph{Train/dev/test splits} For the PE corpus, we use the same split into training and test data as in \newcite{Stab:2017}. In particular, our test data comprises 80 documents (``essays'') with a total of 29,537 tokens (en). We choose 10\% of the training data as dev set. Thus, we have 286 essays in the train set with a total of 105,988 tokens (en) and 36 essays in the dev set with a total of 12,657 tokens (en). 
We report averages over five random initializations of our networks.
For the CRC corpus, we perform 5-fold cross-validation on document level. Our train sets consist of roughly 15K tokens, our dev sets of 2K tokens, and our test sets of 4K tokens. For each split, we average over five different random initializations and report the average over these averages.
For MTX, we also perform 5-fold cross-validation on document level. Our train sets consist of roughly 6K tokens (en), our dev sets of 500 tokens (en), and our test sets of 1,500 tokens (en). We use the same averaging strategy as for CRC, but average over ten random initializations per fold, to account for the smaller dataset sizes.


%% file: experiments.tex
\subsection{Results}
We report results for adapting between datasets in different languages and between parallel versions of one and the same dataset. We only consider cases where one of the involved languages is English. Further, we do not transfer between MTX and the other datasets, because MTX has no ``O'' units (and no major claims). 
Unless stated otherwise, we always \emph{evaluate} on \texttt{HT} for both direct transfer and projection. 
\subsubsection{Direct Transfer}
For all cross-lingual direct transfer experiments, we train on the union of train and dev (train$+$dev) sets (randomly drawn for the datasets for which we used cross-validation) of the source language and test either on the whole data (train$+$dev$+$test) of L2, or, in case of parallel versions of a dataset (such as PE EN$\leftrightarrow$DE) on the test set of L2. We do not use  \texttt{MT} for direct transfer at any stage. 

\textbf{
PE$_{\text{EN}}$$\leftrightarrow$PE$_{\text{DE}}$}, results in Table \ref{table:plaindt}: 
English in-language results do not vary much and are on a level of slightly above 
69\% macro-F1, 
largely independent of the embedding types and whether or not 
character information is available.\footnote{Our in-language results are slightly below our previous results reported in \newcite{eger-daxenberger-gurevych:2017:Long} (table 6), where we obtained scores of 72-75\% for token-level component extraction, even though the architecture is in principle the same. Reasons may be the different word embeddings used as well as that we reported majority performance over different hyperparameter combinations in the previous work, which typically increased performance scores by a few percentage points.}
German in-language results are 4-5 percentage points (pp) below the English ones. One might suspect the presumed inferior quality (or derivative nature) of the student translations as a cause for this, but we hypothesize that German is simply more complex than English, both in morphology and syntax. 

We observe a noticeable drop when moving cross-language. This drop is up to 
$>$40\% for the direction EN$\rightarrow$DE (worst case drop from $\sim$70\% to $\sim$37\%) and slightly less for DE$\rightarrow$EN. We explain this drop by the discrepancy between 
training and test distributions. 
This discrepancy is present even in bilingual embedding spaces: no test word has the exact same representation as the words in the training data. 
Further, disabling character information 
typically has a very positive impact cross-language. For example, EN$\rightarrow$DE performance increases from $\sim$40\% F1 to $\sim$50\% when disabling character information. The reason is that a system that extracts a character representation based on English characters may get confused from the diverging German character sequences. Surprisingly, characters 
do not impede so much in the direction DE$\rightarrow$EN.
The reason seems to be 
lexical borrowing in modern German from English. For example, $\sim$17\% of the `active' vocabulary (i.e., frequency $>$30) of English in  PE$_{\text{EN}}$ 
is also contained in PE$_{\text{DE}}$. In contrast, only 6\% of the active vocabulary of German in PE$_{\text{DE}}$ occurs also in PE$_{\text{EN}}$. 

\textbf{CRC$\leftrightarrow$PE$_{\text{EN}}$}, results in Table \ref{table:micro_crc} (left): In-language CRC results are lower than in-language PE results ($\sim$46\% vs.\  $\sim$69\% for PE). This is unsurprising since CRC is considerably smaller in size than PE. However, we observe that the cross-language drop is much larger than it is for the PE DE$\leftrightarrow$EN setting. In fact, performance values always lie below our random baseline. We attribute this huge drop 
not to the larger language distance between English and Chinese (relative to English and German), but primarily to the domain gap between student essays and hotel reviews. In fact, we observe that, e.g., major claims in PE are almost always preceded by specific discourse markers such as ``Therefore, I believe that'' or ``In the end, I think'' (cf. Table \ref{table:example}), while hotel reviews completely lack such discourse connectives (cf.\ Table \ref{table:example-crc}). Since we expect a system that trains on PE to learn the signaling value of these markers, directly applying this system to text where such markers are absent, fails.%

\textbf{MTX$_{\text{EN}}$$\leftrightarrow$MTX$_{\text{DE}}$}, results in Table \ref{table:micro_crc} (right): 
Even though the dataset is by far smallest in size it yields the highest F1-scores among all our considered datasets. Moreover, the language drop is 
comparatively small 
(between 4 and 7pp). 
Investigating, 
we notice that 
argument components are typically separated by punctuation symbols (mostly ``.'' or ``,'') in MTX, which is easy to learn even cross-lingually. Moreover, we find that claims can often be separated from premises by simple keywords such as ``should'', %
which can, apparently, be reliably spotted cross-lingually via the corresponding bilingual word embeddings.  

\begin{table*}[!htb]
  \centering
  { \small
\begin{tabular}{lc|cccc} \toprule
Model & Embedding Type & EN$\rightarrow$EN & EN$\rightarrow$DE &
DE$\rightarrow$DE & DE$\rightarrow$EN \\ \midrule
BLCRF$+$Char & BIVCD-100 & {68.87} & 41.89 & 65.22 & 49.91 \\
            & BIVCD-200 & \textbf{70.51} & 39.87 & \textbf{65.92} & {49.52}  \\
            & BISKIP-100 & {69.27} & 37.01 & 63.33 & 48.23 \\
BLCRF        & BIVCD-100 & {69.27} & 49.70 & 65.90 & 50.14  \\
            & BISKIP-100 & 69.15 & \textbf{49.76} & {64.92} & \textbf{50.28}  \\ \midrule
Baseline & & 20. & 20. & 20. & 20.  \\
\bottomrule
\end{tabular}
}
\caption{Direct transfer results for PE$_\text{EN}$$\leftrightarrow$PE$_\text{DE}$. Scores are macro-F1.} 
\label{table:plaindt}
\end{table*}

\begin{table*}[!ht]
  \centering
  \small
  \begin{tabular}{l|rrrr|rrrr} \toprule
  & 
 \multicolumn{4}{c}{CRC$\leftrightarrow$PE$_{\text{EN}}$} & \multicolumn{4}{c}{MTX$_\text{EN}$$\leftrightarrow$MTX$_{\text{DE}}$}\\
     Model & ZH$\rightarrow$ZH & ZH$\rightarrow$EN & EN$\rightarrow$EN & EN$\rightarrow$ZH & EN$\rightarrow$EN & EN$\rightarrow$DE & DE$\rightarrow$DE & DE$\rightarrow$EN
    \\ \midrule
    BLCRF$+$Char  &  \textbf{46.31} & 14.01  & \textbf{69.27} & 9.50
    & \textbf{73.12} & 67.03  & \textbf{73.41} & \textbf{66.62}
    \\
    BLCRF & 44.95 & {16.52} & {69.15} & {12.60}
    & 72.15 & \textbf{69.46} & 72.52 & 63.71 
    \\ \midrule
    Baseline & 18. & \textbf{17.} & 20. & \textbf{17.}
    & 45. & 46. & 50. & 50. 
    \\
    \bottomrule
  \end{tabular}
  \caption{{Direct transfer results for CRC and MTX. Scores are macro-F1. Embeddings are BISKIP-100.}}
  \label{table:micro_crc}
\end{table*}

\paragraph{Error Analysis and Discussion}
For PE direct transfer experiments, we find that 
a major source of errors is incorrect classification of 
tokens labeled ``B-''. This means that the system has difficulty finding the exact beginning of an argument span. 
We find, however, that the reason for the 
language drop is not that the bilingual embedding spaces are bad: among the top-10 neighbors of English words are roughly five German words, and vice versa. 
Rather, direct transfer induces a situation very similar to standard monolingual out-of-vocabulary (OOV) scenarios, namely as if all test words had been replaced by synonyms that did not occur in the train data. While systems using embeddings as input are more robust to OOV words, they 
are still affected by them \cite{Ma:2016,Mueller:2013}. 
This ``blurring effect'' at test time then makes it more difficult to detect exact argument component spans. 
While this is true in general, it is not true for punctuation symbols, which typically have an identical role across languages and, hence, extremely similar representations. For example, German and English ``.'' have more than 97\% cosine similarity in BISKIP-100d, which is much higher than for 
typical monolingually closely related words. 
Finally, besides semantic shift direct transfer also faces syntactic shift, because the test words may have different word order compared to the train data (e.g., verb final position in German).

The lessons we learn from our above experiments are that (i) the MTX dataset does not provide a real challenge for cross-lingual techniques because argument components can easily be spotted based on punctuation and component typing appears to be just as easily portable across languages. (ii) Language adaptation between the CRC corpus and PE appears, in contrast, too difficult because argumentation units are very differently realized across the two datasets, and hence, the domain shift appears to be the (much) larger obstacle compared to the language shift.\footnote{This is a similar finding as in \newcite{Daxenberger:2017}.} Thus, (iii) we mostly focus on the cross-lingual version of the PE corpus in the sequel, which is a difficult enough dataset for cross-lingual AM, without confounding the problem with issues relating to differences of AM domains. In addition, we only use BISKIP-100d embeddings, because the choice of embeddings seemed to have a negligible effect in our case (certainly, for our main focus, namely, cross-lingual evaluations) and because they showed slightly superior results cross-lingually than BIVCD-100d.

\subsubsection{Projection}
\paragraph{\texttt{HT}-Projection}
Table~\ref{table:projection} (\texttt{HT} columns) shows results when we project PE$_\text{EN}$ training data annotations on parallel German \texttt{HT} documents and train and evaluate a system directly on German (and vice versa). As said before, we train in this case directly on the same language as we test on, viz., L2. We observe that this improves cross-language results dramatically. From a best cross-language result of 49.76\% for BIVEC-100d in the direct transfer setting, we improve by almost 30\% to 63.67\%. This is only roughly 1pp below the best in-language result for German which was 64.92\%. In the direction DE$\rightarrow$EN, we observe the same trend: we improve by over 30\% relative to the direct transfer results and achieve a macro-F1 score that is only 1.7pp below the in-language upper-bound.  

\paragraph{\texttt{MT}-Projection}
Next, we investigate what happens when we replace the \texttt{HT} translations with \texttt{MT} translations and perform the same projection steps as before. Results for PE$_\text{EN}$$\leftrightarrow$PE$_\text{DE}$ are shown in Table \ref{table:projection} (\texttt{MT} columns). We see that EN$\rightarrow$DE results get slightly better while DE$\rightarrow$EN results get slightly worse. On average, it seems, using machine translations is just as good as using human translations. Moreover, we remain very close to 
the upper-bound in-language results. The reason why \texttt{MT} results could be better than \texttt{HT} results is that the machine might translate more consistently. It might also be better in certain cases in correcting (the sometimes ungrammatical) English original. Another likely reason is that current \texttt{MT} has reached, if not already surpassed, \texttt{HT} of non-expert (but bilingually fluent) human translators.\footnote{We conducted a formal test if \texttt{MT} can reliably be distinguished from \texttt{HT} in our setup. We trained a system (an adaptation of InferSent \cite{Conneau:2017}) to predict whether for an English original $e$ and a second input sentence it could determine if the second input is a human or machine translation of $e$. The system's performance of 54\% accuracy (which is only slightly better than random guessing) matched our own intuition and introspection into the quality of the machine translations.}

Motivated by this finding, we also machine translated the CRC corpus into English, projected annotations and then trained a system on this translation and evaluated on the PE$_{\text{EN}}$ corpus. 
Results improve to 23.15\% macro-F1 score relative to the best direct transfer result of 16.52\%. This indicates, again, that training in L2 is better than training in L1, given the high quality machine translations and a suitable projection algorithm. 
However, 23.15\% is still only slightly better than the random baseline of 17\%---and far from the best PE$_{\text{EN}}$ in-language result of $>$69\%. 
In addition, in the direction PE$_\text{EN}$$\rightarrow$CRC macro-F1 (also) improves to 15.33\% relative to a best direct transfer score of 12.60\%. However, here the performance is still below the random baseline of 17\%. 
We take this as strong evidence that the domain gap between CRC and PE is too large and it is not possible to train a system in one of these two domains and directly apply it in the other, even when the language gap has been eliminated.  

\begin{table}[!htb]
 \centering
 \small
  \begin{tabular}{lccc|ccc} \toprule
  	  & \multicolumn{3}{c|}{EN$\rightarrow$DE} & \multicolumn{3}{c}{DE$\rightarrow$EN} \\ 
      &\texttt{HT}&\texttt{MT}&In-Lang.&\texttt{HT}&\texttt{MT}&In-Lang.\\ \midrule
     BLCRF$+$Char  & 63.67 & \bf 64.00 & 63.33 & \bf 67.57 & 66.39 & 69.27 \\
     BLCRF & 61.18 & 63.34 & 64.92 & 64.87 & 64.68 & 69.15\\ 
    \bottomrule
  \end{tabular}
  \caption{Projection on \texttt{HT}/\texttt{MT} translations, evaluated on human-created test data. Scores are macro-F1. Embeddings are BISKIP-100.}
  \label{table:projection}
\end{table}

\paragraph{Other languages} We conducted a final experiment in which we considered our \texttt{MT} translations of PE into 
French, Spanish, and Chinese. Since we have no human-created test data for these languages we could only evaluate on \emph{machine translations {and} projected annotations}. 
For our 
BLCRF$+$char model, 
we obtained performance scores of 62.45\%, 65.92\%, 59.20\% for French, Spanish and Chinese, respectively. 
To see if these numbers give reliable estimates of the systems when evaluated on \texttt{HT} data, we performed the same test with German and English 
and got scores of 63.20\% and 61.45\%, respectively, when trained and evaluated on \texttt{MT}. 
For CRC, we also trained and evaluated on the English (\texttt{MT} translated and automatically projected) data, obtaining a score of 47.92\% with BLCRF$+$char, which is slightly above the in-language value of 46.31\% 
on the original Chinese data (see Table \ref{table:micro_crc}). 
That all these numbers are close to the original in-language results gives a good indication that the \texttt{MT} evaluations very likely strongly correlate to `true' performances on \texttt{HT} data.

\paragraph{Error analysis}
The bottleneck of projection is the quality of the cross-lingual projections (which in turn depend on the quality of the word alignments between bi-text). We can directly assess 
our projections by comparing them to the human-created annotations. 
Our algorithmic projections match in 97.24\% of the cases (token-level) with the human gold standard for the direction EN$\rightarrow$DE. The corresponding macro-F1 score is 89.85\%. Inspecting the confusion matrices, we observe that most mismatches occur between the ``B'' and ``I'' categories of a given component type and with the ``O'' category. 
These numbers and the mismatch types indicate that 
there are only few projection errors and they typically lead to 
either (slightly) larger or smaller argument components than given in the human-created data. 
To illustrate, typical projection errors arise in case of missing articles in one of the two languages involved; cf.~Table~\ref{table:example}: ``luck is [...]'' vs.\ ``la chance est [...]''. 
Here, it is likely that the alignment algorithm does not align the French determiner ``la'' to an English word, and thus, ``la'' is not included in the argument component. 
Another case of too short argument components is that of verb final position in German which often gets misaligned and the corresponding final verb omitted from the argument component. 
These misalignments lead to slightly ``shifted'' argument components in the L2 train set and are the most prominent source of error for the projection technique. 
To quantify: the German in-language system classifies 58\% of all cases when one of the determiners (``der/die/das'') begins an argument component correctly, but the system using projections from the English data only classifies 35\% of them correctly. Hence, alignment/projection errors indeed propagate to some degree, 
but these phenomena are rare and have negligible impact on performance. Note that, by design of our projection algorithm, misalignments of words in the `center' of an argument component are much less likely to be a problem. 

\begin{figure}
\centering
\scalebox{0.5}[0.5]{\input{plots/individual_f1_8}}
\caption{Individual F1-scores for four indicated systems and seven labels. All transfer systems are from PE$_{\text{DE}}$ to PE$_{\text{EN}}$; EN is in-language. DT stands for Direct Transfer. \texttt{HT}/\texttt{MT} are projection-based approaches. Embeddings are BISKIP-100. Systems are BLCRF$+$Char.
}
\label{fig:f1}
\end{figure}
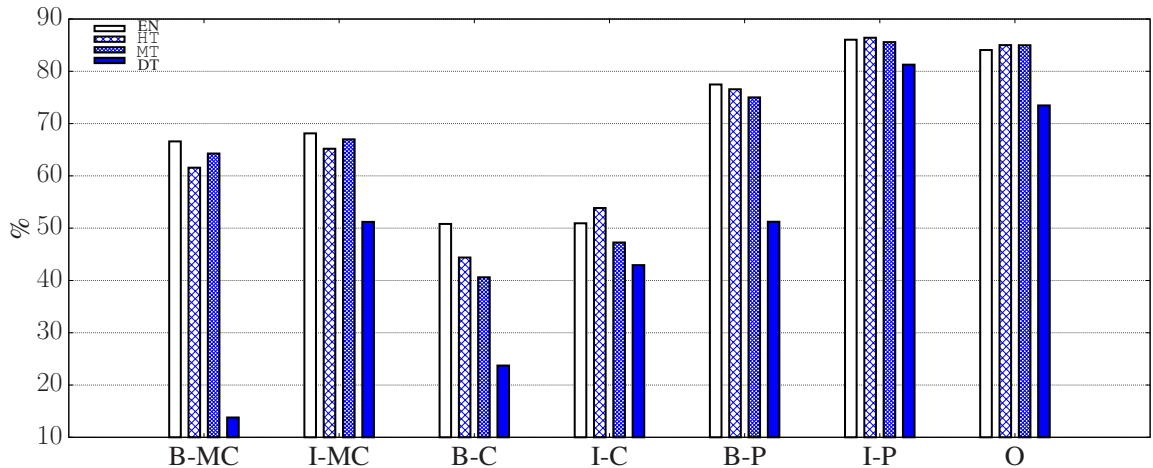

Figure \ref{fig:f1} plots individual F1 scores for various systems transferring to English on PE. Here, the cross-lingual systems using \texttt{HT/MT} projection perform roughly as well as the in-language system for ``O'',  ``I-C'' and ``I-P''. These are the most frequent classes. For claims and major claims, which have lower frequency, the in-language upper bound tends to perform better. Noteworthy, the in-language system is always better for the beginnings of an argument component (B-MC,B-C,B-P), which confirms our above analysis. The direct transfer system, in contrast, performs much worse, particularly for major claims and claims, and also for all starts of components, indicating that the blurring (``OOV'') effect is here much more severe. 

%% file: plots/individual_f1_8.tex
\begingroup
  \makeatletter
  \providecommand\color[2][]{%
    \GenericError{(gnuplot) \space\space\space\@spaces}{%
      Package color not loaded in conjunction with
      terminal option `colourtext'%
    }{See the gnuplot documentation for explanation.%
    }{Either use 'blacktext' in gnuplot or load the package
      color.sty in LaTeX.}%
    \renewcommand\color[2][]{}%
  }%
  \providecommand\includegraphics[2][]{%
    \GenericError{(gnuplot) \space\space\space\@spaces}{%
      Package graphicx or graphics not loaded%
    }{See the gnuplot documentation for explanation.%
    }{The gnuplot epslatex terminal needs graphicx.sty or graphics.sty.}%
    \renewcommand\includegraphics[2][]{}%
  }%
  \providecommand\rotatebox[2]{#2}%
  \@ifundefined{ifGPcolor}{%
    \newif\ifGPcolor
    \GPcolortrue
  }{}%
  \@ifundefined{ifGPblacktext}{%
    \newif\ifGPblacktext
    \GPblacktexttrue
  }{}%
  \let\gplgaddtomacro\g@addto@macro
  \gdef\gplbacktext{}%
  \gdef\gplfronttext{}%
  \makeatother
  \ifGPblacktext
    \def\colorrgb#1{}%
    \def\colorgray#1{}%
  \else
    \ifGPcolor
      \def\colorrgb#1{\color[rgb]{#1}}%
      \def\colorgray#1{\color[gray]{#1}}%
      \expandafter\def\csname LTw\endcsname{\color{white}}%
      \expandafter\def\csname LTb\endcsname{\color{black}}%
      \expandafter\def\csname LTa\endcsname{\color{black}}%
      \expandafter\def\csname LT0\endcsname{\color[rgb]{1,0,0}}%
      \expandafter\def\csname LT1\endcsname{\color[rgb]{0,1,0}}%
      \expandafter\def\csname LT2\endcsname{\color[rgb]{0,0,1}}%
      \expandafter\def\csname LT3\endcsname{\color[rgb]{1,0,1}}%
      \expandafter\def\csname LT4\endcsname{\color[rgb]{0,1,1}}%
      \expandafter\def\csname LT5\endcsname{\color[rgb]{1,1,0}}%
      \expandafter\def\csname LT6\endcsname{\color[rgb]{0,0,0}}%
      \expandafter\def\csname LT7\endcsname{\color[rgb]{1,0.3,0}}%
      \expandafter\def\csname LT8\endcsname{\color[rgb]{0.5,0.5,0.5}}%
    \else
      \def\colorrgb#1{\color{black}}%
      \def\colorgray#1{\color[gray]{#1}}%
      \expandafter\def\csname LTw\endcsname{\color{white}}%
      \expandafter\def\csname LTb\endcsname{\color{black}}%
      \expandafter\def\csname LTa\endcsname{\color{black}}%
      \expandafter\def\csname LT0\endcsname{\color{black}}%
      \expandafter\def\csname LT1\endcsname{\color{black}}%
      \expandafter\def\csname LT2\endcsname{\color{black}}%
      \expandafter\def\csname LT3\endcsname{\color{black}}%
      \expandafter\def\csname LT4\endcsname{\color{black}}%
      \expandafter\def\csname LT5\endcsname{\color{black}}%
      \expandafter\def\csname LT6\endcsname{\color{black}}%
      \expandafter\def\csname LT7\endcsname{\color{black}}%
      \expandafter\def\csname LT8\endcsname{\color{black}}%
    \fi
  \fi
    \setlength{\unitlength}{0.0500bp}%
    \ifx\gptboxheight\undefined%
      \newlength{\gptboxheight}%
      \newlength{\gptboxwidth}%
      \newsavebox{\gptboxtext}%
    \fi%
    \setlength{\fboxrule}{0.5pt}%
    \setlength{\fboxsep}{1pt}%
\begin{picture}(17006.00,6802.00)%
    \gplgaddtomacro\gplbacktext{%
      \csname LTb\endcsname%
      \put(496,320){\makebox(0,0)[r]{\strut{}\huge$10$}}%
      \csname LTb\endcsname%
      \put(496,1106){\makebox(0,0)[r]{\strut{}\huge$20$}}%
      \csname LTb\endcsname%
      \put(496,1892){\makebox(0,0)[r]{\strut{}\huge$30$}}%
      \csname LTb\endcsname%
      \put(496,2678){\makebox(0,0)[r]{\strut{}\huge$40$}}%
      \csname LTb\endcsname%
      \put(496,3465){\makebox(0,0)[r]{\strut{}\huge$50$}}%
      \csname LTb\endcsname%
      \put(496,4251){\makebox(0,0)[r]{\strut{}\huge$60$}}%
      \csname LTb\endcsname%
      \put(496,5037){\makebox(0,0)[r]{\strut{}\huge$70$}}%
      \csname LTb\endcsname%
      \put(496,5823){\makebox(0,0)[r]{\strut{}\huge$80$}}%
      \csname LTb\endcsname%
      \put(496,6609){\makebox(0,0)[r]{\strut{}\huge$90$}}%
      \put(2608,0){\makebox(0,0){\strut{}{\huge B-MC}}}%
      \put(4623,0){\makebox(0,0){\strut{}{\huge I-MC}}}%
      \put(6639,0){\makebox(0,0){\strut{}{\huge B-C}}}%
      \put(8655,0){\makebox(0,0){\strut{}{\huge I-C}}}%
      \put(10670,0){\makebox(0,0){\strut{}{\huge B-P}}}%
      \put(12686,0){\makebox(0,0){\strut{}{\huge I-P}}}%
      \put(14701,0){\makebox(0,0){\strut{}{\huge O}}}%
    }%
    \gplgaddtomacro\gplfronttext{%
      \csname LTb\endcsname%
      \put(-100,3464){\rotatebox{-270}{\makebox(0,0){\strut{}\huge\%}}}%
      
      \csname LTb\endcsname%
      \put(1940,6490){\makebox(0,0)[r]{\strut{} \large EN}}%
      \csname LTb\endcsname%
      \put(1950,6306){\makebox(0,0)[r]{\strut{} \Large \texttt{HT}}}%
      \csname LTb\endcsname%
      \put(1950,6100){\makebox(0,0)[r]{\strut{} \Large \texttt{MT}}}%
      \csname LTb\endcsname%
      \put(1940,5900){\makebox(0,0)[r]{\strut{} \large DT}}%
    }%
    \gplbacktext
    \put(0,0){\includegraphics{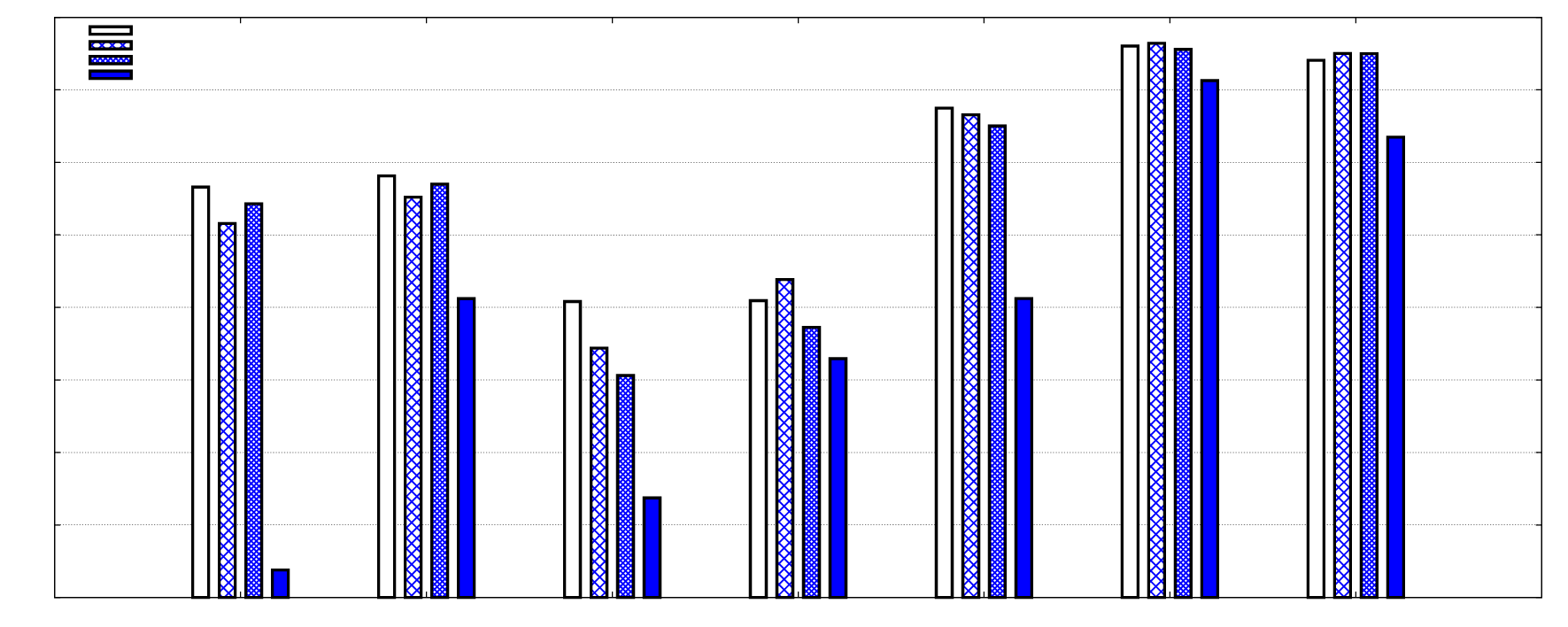}}%
    \gplfronttext
  \end{picture}%
\endgroup

%% file: conclusion.tex
Showing that the currently available datasets for AM are not adequate for evaluating cross-lingual AM transfer, we created human and machine translations of one of the most popular current AM datasets, the dataset of persuasive student essays \cite{Stab:2017}. 
We also machine translated a Chinese corpus of reviews \cite{Li2017} into English, which provides argumentation structures on hotel reviews. 
Performing cross-lingual experiments using suitable adaptations of two popular transfer approaches, we have shown that machine translation and (naive) projection work considerably better than direct transfer, even though the former approach contains two sources of noise. Moreover, machine translation in combination with projection almost performs on the level of in-language upper bound results. 
We think that our findings shed further light on the value---and the huge potential---of current (neural) machine translation systems for cross-lingual transfer. {They also cast doubt on current standard use of direct transfer in cross-lingual scenarios. Instead, we propose to simply machine translate the train set, when this is possible, and then project labels to the translated text. This eliminates the (particular) ``OOV'' and ``ordering'' problems inherent to direct transfer. Prerequiste to this approach is high quality MT, which, with the advent of neural techniques, appears to be now available.} 

We hope our new datasets fuel AM research in languages other than English. 
In this work, we did not consider cross-lingual argumentative relation identification, although relations are available in the newly created parallel PE and CRC datasets.
Future work should explore cross-lingual multi-task learning for AM \cite{Schulz:2018:NAACL} with the source language as main task and small amounts of labeled target language data, as well as adversarial training techniques \cite{Yasunaga:2018}, which promise to be beneficial for the particular OOV problem that direct transfer is prone to (though not for the ordering problem).
We also want to combine projection with direct transfer by training on the union of projected L2 data as well as the original L1 data using shared representations. 